\documentclass[11pt]{article}
\usepackage{authblk}

\newlength{\defbaselineskip}
\usepackage{graphicx}
\usepackage{amsmath}
\usepackage{amssymb}
\usepackage[caption=false,font=footnotesize]{subfig}

\usepackage{algorithmic}
\usepackage{algorithm}

\begin{document}

\title{
Large-Scale Sparse Subspace Clustering Using Landmarks\footnote{This paper is accepted for publication in 2019 IEEE International Workshop on Machine Learning for Signal Processing (MLSP), Pittsburgh, PA, USA.
}}

\author{ \Large Farhad Pourkamali-Anaraki}
\date{ Department of Computer Science, University of Massachusetts Lowell, MA, USA}


\maketitle

\begin{abstract}
	Subspace clustering methods based on expressing each data point as a linear combination of all other points in a dataset are popular unsupervised learning techniques. However, existing methods incur high computational complexity on large-scale datasets as they require solving an expensive optimization problem and performing spectral clustering on large affinity matrices. This paper presents an efficient approach to subspace clustering by selecting a small subset of the input data called landmarks. The resulting subspace clustering method in the reduced domain runs in linear time with respect to the size of the original data. Numerical experiments on synthetic and real data demonstrate the effectiveness of our method.
\end{abstract}

\section{Introduction}
\label{sec:intro}
A fundamental task in machine learning and modern data analysis is to infer the latent structure underlying unlabeled datasets that are composed of a mixture of several components \cite{TowardsKmeans}. For example, digital pathology images are represented by high-dimensional vector spaces  corresponding to the number of pixels. However, a few relevant features  are often sufficient to identify groups of similar images \cite{peikari2018cluster}. Therefore, a flexible approach for modeling heterogeneous datasets focuses on the assumption that data points are drawn from a union of unknown subspaces, where the dimension of each subspace describes the number of degrees of freedom. This problem is known as subspace clustering (SC), which consists of partitioning the data into multiple clusters and fitting each cluster with a low-dimensional subspace \cite{VidalSubspace,soltanolkotabi2012geometric,soltanolkotabi2014robust,mcwilliams2014subspace}. 

A number of approaches to subspace clustering have been proposed in the literature such as algebraic, statistical, and spectral clustering methods; see \cite{elhamifar2013sparse} for a review. Among existing techniques, SC methods based on the self-expressiveness property \cite{elhamifar2009sparse} are popular due to their great empirical success and theoretical guarantees. This line of work, referred to as sparse subspace clustering (SSC), follows a two-step approach. The first step involves representing each data point as a linear combination of all other data points to build an affinity matrix of a graph with vertices corresponding to data points. The second step of SSC  applies normalized spectral clustering \cite{von2007tutoria}  to the learned affinity matrix.

While SSC has drawn significant attention over the past few years, the computational complexity of these methods has to date stood as an obstacle when analyzing large datasets. The construction of affinity matrices incurs high computational cost as one should search for a regularized representation of every single data point with respect to the whole dataset. Moreover, spectral clustering is computationally expensive on large-scale datasets because it involves an eigenvalue decomposition. A straightforward implementation of spectral clustering has a cubic time complexity with respect to the number of data points. However, there are approximate algorithms which run in quadratic time with respect to the size of data \cite{yan2009fast,tian2014learning,pourkamali2016randomized,chen2018spectral}. As a result, the overall computational complexity (quadratic or cubic) in the input size renders existing SSC algorithms  expensive or even prohibitive in the presence of large-scale data.

This paper presents a new method for reducing the computational burden of SSC when the number of data points is large. Given a dataset, the proposed method first selects a small subset of the input data called landmarks, and then represents each data point as a linear combination of the landmarks (rather than the entire dataset as in SSC). The underlying idea is that the required number of data points for solving the SSC problem is proportional to the number of subspaces and their intrinsic dimensions \cite{soltanolkotabi2014robust}. In this paper, we consider two efficient landmark selection strategies: (1) uniform sampling, where a  subset of the input data is selected uniformly at random; and (2) nonuniform sampling scheme by  obtaining a rough approximation of partitioning the input data using K-medoids clustering \cite{park2009simple}. The latter is more suitable to select representative points from all clusters present in the dataset.

The second contribution  is to show that the spectral embedding of the input data  can be obtained efficiently without the need to explicitly form affinity matrices and compute their eigenvalue decompositions. In fact, the computational cost of spectral clustering in our proposed method scales linearly with the data size when the number of landmarks is fixed. Thus, the use of landmarks in our proposed method allows us to simultaneously reduce the computational cost associated with both steps of SSC. Whereas previous methods primarily attempt to reduce the computational cost of the first step, and rely heavily on forcing affinity matrices to be sparse.  

The rest of this paper is organized as follows. Section \ref{sec:prelim} provides the SSC problem formulation and a brief overview of prior art. Section \ref{sec:method} introduces the proposed  SSC method for large-scale data using landmarks. Extensive experiments in Section \ref{sec:illust} demonstrate the effectiveness and computational benefits of our method in comparison with existing algorithms, while Section \ref{sec:conc} provides concluding remarks. 

\section{Preliminaries}
\label{sec:prelim}
Consider  a set of $n$ data points $\mathcal{X}=\{\mathbf{x}_1,\ldots,\mathbf{x}_n\}\subseteq \mathbb{R}^D$ that are drawn from a union of $K$ subspaces $\{\mathcal{S}_k\}_{k=1}^K$ as follows
\begin{align}
\mathbf{x}_i=\mathbf{H}_k\mathbf{z}_i+\mathbf{e}_i,\;\forall \mathbf{x}_i\in\mathcal{S}_k, \label{eq:data}
\end{align}
where the columns of $\mathbf{H}_k\in\mathbb{R}^{D\times d_k}$ form an orthonormal basis of $\mathcal{S}_k$. Thus, $d_k<D$ is the dimensionality of $\mathcal{S}_k$, $\mathbf{z}_i\in\mathbb{R}^{d_k}$ is the low-dimensional representation of $\mathbf{x}_i$ with respect to $\mathcal{S}_k$, and  $\mathbf{e}_i\in\mathbb{R}^D$ is the noise vector. Given the unlabeled dataset, subspace clustering (SC) aims to recover the underlying subspaces  and the segmentation of  points according to the subspaces. Once   data-to-subspace assignments are obtained, one can recover the subspaces $\{\mathcal{S}_k\}_{k=1}^K$ using the singular value decomposition on the data associated with each cluster \cite{traganitis2018sketched}. 

\subsection{SSC problem formulation}\label{sec:form}
A popular approach to solve the SC problem is to form an affinity matrix based on expressing each data point as a linear combination of other points from its own subspace. That is, we would like to express each data point $\mathbf{x}_j$ in the form of $\mathbf{x}_j=\sum_{i\neq j}c_{ij}\mathbf{x}_i$ such that the coefficient $c_{ij}$ is nonzero when $\mathbf{x}_j$ and $\mathbf{x}_i$ are from the same subspace. Note that if the subspace dimensions $d_k$, $k=1,\ldots,K$, are small compared to $D$, then these coefficients are sparse. Thus, sparse subspace clustering (SSC) solves the following optimization problem
\begin{align}
\min_{\mathbf{c}_j\in\mathbb{R}^n} \|\mathbf{c}_j\|_1+\frac{\lambda}{2} \|\mathbf{x}_j - \sum_{i\neq j}c_{ij}\mathbf{x}_i\|_2^2,\;\forall \mathbf{x}_j\in\mathcal{X},\label{eq:SSC}
\end{align}
where $\|\cdot\|_q$ denotes the $\ell_q$ norm for vectors, $\lambda>0$ is the regularization parameter, and $\mathbf{c}_j=[c_{1j},\ldots,c_{nj}]^T$. Therefore, SSC searches for a coefficient matrix $\mathbf{C}=[\mathbf{c}_1,\ldots,\mathbf{c}_n]\in\mathbb{R}^{n\times n}$ such that $\mathbf{X}\approx \mathbf{X}\mathbf{C}$, where $\mathbf{X}=[\mathbf{x}_1,\ldots,\mathbf{x}_n]\in\mathbb{R}^{D\times n}$ is the data matrix. Using the encoded information about the membership of data points to the subspaces in $\mathbf{C}$, we form a graph with $n$ vertices and the affinity matrix $\mathbf{W}=|\mathbf{C}|+|\mathbf{C}^T|$.

In the second step of SSC, the segmentation of data is found by applying spectral clustering to $\mathbf{W}$. To this end, we form the normalized graph  Laplacian matrix $\mathbf{L}=\mathbf{I} - \mathbf{D}^{-1/2}\mathbf{W}\mathbf{D}^{-1/2}$, where $\mathbf{D}$ is the diagonal degree matrix such that its $i$-th diagonal element is equal to $\sum_{j=1}^{n}w_{ij}$ and $\mathbf{I}$ is the identity matrix. Then, the $K$ eigenvectors $\mathbf{v}_1,\ldots,\mathbf{v}_K\in\mathbb{R}^n$ corresponding to the $K$ smallest eigenvalues of $\mathbf{L}$, or equivalently the $K$ largest eigenvalues of $\mathbf{D}^{-1/2}\mathbf{W}\mathbf{D}^{-1/2}$, are found. Finally, K-means clustering \cite{pourkamali2017preconditioned} is performed on the rows of the matrix $\mathbf{V}=[\mathbf{v}_1,\ldots,\mathbf{v}_K]\in\mathbb{R}^{n\times K}$ to obtain  the data-to-subspace assignments for the $n$ input data points. 

\subsection{Prior work}\label{sec:prior}
Previous work on developing efficient SSC methods can be divided into two categories. The first line of work aims to reduce the computational cost associated with solving the optimization problem in \eqref{eq:SSC} using the whole dataset. For example, the authors in \cite{elhamifar2013sparse} proposed to use the  alternating direction method of multipliers (ADMM) which takes $O(n^3)$ flops \cite{boyd2011distributed}
\begin{align}
\min_{\mathbf{C}} \|\mathbf{C}\|_1+\frac{\lambda}{2}\|\mathbf{X}-\mathbf{X}\mathbf{C}\|_F^2\;\text{s.t.}\;\text{diag}(\mathbf{C})=\mathbf{0},\label{eq:ADMM}
\end{align}
where $\|\cdot\|_1$ and $\|\cdot\|_F$ represent the matrix $\ell_1$ norm and the Frobenius norm, respectively. Also, $\text{diag}(\mathbf{C})$ is a vector that contains the diagonal elements of $\mathbf{C}$ so that we avoid the trivial solution of \eqref{eq:ADMM}. Recently, a more efficient implementation of ADMM for SSC was  introduced with a cost that is quadratic in $n$ \cite{pourkamali2018efficient}. Furthermore, the authors in \cite{you2016scalable} proposed to use orthogonal matching pursuit (OMP) for finding sparse representations  by enforcing a strict $\ell_0$ sparsity constraint instead of the $\ell_1$ norm. The resulting SSC-OMP algorithm is more efficient than ADMM (roughly quadratic in $n$ depending on the specific implementation). It is worth noting that  the described methods mainly focus on the construction of affinity matrices, and the spectral clustering step  still involves  eigenvalue decompositions of $n\times n$ matrices.

A recent work \cite{you2018scalable} takes a different approach by  selecting a small subset of $\mathcal{X}$ that represents all data points. Each  point in the dataset is then expressed as a linear combination of points in the selected subset $\mathcal{X}_0\subseteq\mathcal{X}$ of size $m<n$ 
\begin{align}
\min_{\mathbf{c}_j\in\mathbb{R}^m} \|\mathbf{c}_j\|_1+\frac{\lambda}{2}\|\mathbf{x}_j - \sum_{i: \mathbf{x}_i\in\mathcal{X}_0}c_{ij}\mathbf{x}_i\|_2^2,\;\forall \mathbf{x}_j\in\mathcal{X}.\label{eq:EX}
\end{align}
The authors in \cite{you2018scalable} also proposed a new method for selecting the subset $\mathcal{X}_0$ based on minimizing a maximum representation cost of $\mathcal{X}$ (called ESC), and solving the above optimization problem has a linear time complexity in $n$. Geometrically,  a subset of data points is selected to best cover the entire dataset as measured by the Minkowski functional. In this approach, the coefficient matrix $\mathbf{C}=[\mathbf{c}_1,\ldots,\mathbf{c}_n]$ is $m\times n$ and, thus,  spectral clustering cannot be directly applied to $\mathbf{W}$ as explained in Section \ref{sec:form}. To fix this problem, for each $\mathbf{c}_j$, the $t$ nearest neighbors with the largest positive inner products are found, and the corresponding entries of $\mathbf{W}\in\mathbb{R}^{n\times n}$ are set to be $1$.  Spectral clustering is then applied to $\mathbf{A}=\mathbf{W}+\mathbf{W}^T$. 

\section{The Proposed Method}
\label{sec:method}
In this section, we introduce an efficient method to reduce the computational complexity of  SSC, while maintaining high levels of clustering accuracy. The key idea behind our  method is that, under the union of subspaces model, the  number of samples required for revealing the underlying structure of the data should be proportional to the number of subspaces $K$ and their intrinsic dimensions $d_1,\ldots,d_K$ (and some other factors such as the noise level). Interestingly, recent research efforts have shown that selecting a small subset of the input data results in accurate and efficient learning of mixture models such as K-means clustering \cite{bachem2018scalable}, Gaussian mixture models \cite{lucic2017training2}, and kernel-based clustering \cite{pourkamali18}. Thus, our goal in this paper is to devise sampling schemes for the SSC problem which will enable complexity-performance tradeoffs. 

In this work, we explore two efficient sampling strategies to select a small subset of the original data called landmarks. The first  technique is based on uniform sampling where $m$ points are selected uniformly at random from the set of $n$ input data points $\mathcal{X}$. The main advantage of this method is its negligible computational cost even for very large datasets. 

However, a nonuniform sampling mechanism which adapts to the specific structure of $\mathcal{X}$ is useful  when the number of landmarks $m$ is small compared to the data size. To this end, we propose to run the K-medoids clustering algorithm \cite{park2009simple} on the original data $\mathcal{X}$ to  partition the data into $m$ clusters. Then,  $m$ cluster centers or medoids are selected as landmarks $\mathcal{X}_0$ for solving the SSC problem.
We choose K-medoids clustering over K-means for obtaining a rough partitioning of the data because of two main reasons: (1) K-medoids is more robust to noise and outliers; and (2) K-medoids returns an actual point in each cluster.

After selecting $m$ landmarks $\mathcal{X}_0$, we represent each point in $\mathcal{X}$ as a linear combination of landmarks $\mathcal{X}_0$ by solving
\begin{align}
\min_{\mathbf{c}_j\in\mathbb{R}^m} \|\mathbf{c}_j\|_1+\frac{\lambda}{2}\|\mathbf{x}_j - \sum_{i: \mathbf{x}_i\in\mathcal{X}_0\setminus\{\mathbf{x}_j\}}c_{ij}\mathbf{x}_i\|_2^2,\forall \mathbf{x}_j\in\mathcal{X}. \label{eq:our}
\end{align}
A difference between our optimization problem and the one in \eqref{eq:EX} is that our method does not allow any data point to be expressed as a linear combination of itself since such coefficients will degrade the performance of SSC when $m$ is large. The time complexity of solving the above optimization problem is linear in $n$ when the number of landmarks is fixed. 

The above optimization problem results in a coefficient matrix $\mathbf{C}$ of size $m\times n$, which encodes information about the similarities between $\mathcal{X}_0$ and $\mathcal{X}$. Obviously, it is not feasible to directly build the affinity matrix $\mathbf{W}$  discussed in Section \ref{sec:form} because $\mathbf{C}$ is not $n\times n$ anymore. To resolve this issue, a naive approach is to explicitly form the $n\times n$ matrix $\mathbf{W}=\widetilde{\mathbf{C}}^T\widetilde{\mathbf{C}}$,  where $\widetilde{\mathbf{C}}=|\mathbf{C}|$ and then construct the normalized Laplacian matrix. We present an alternative approach to efficiently obtain the spectral embedding of the input data. Let us consider the columns of $\widetilde{\mathbf{C}} = [\widetilde{\mathbf{c}}_1,\ldots,\widetilde{\mathbf{c}}_n]\in\mathbb{R}^{m\times n}$, and compute the $i$-th element of the degree matrix $\mathbf{D}$ as follows
\begin{align}
\sum_{j=1}^{n}w_{ij}=\sum_{j=1}^{n} \widetilde{\mathbf{c}}_i^T \widetilde{\mathbf{c}}_j=\widetilde{\mathbf{c}}_i^T\sum_{j=1}^{n}\widetilde{\mathbf{c}}_j.
\end{align}
Hence, one can form a vector $\boldsymbol{\alpha}=\sum_{j=1}^{n}\widetilde{\mathbf{c}}_j\in\mathbb{R}^m$ to compute the $i$-th diagonal element of $\mathbf{D}$ as the inner product between $\widetilde{\mathbf{c}}_i$ and $\boldsymbol{\alpha}$ for $i=1,\ldots,n$. Next, we can find the eigenvalue decomposition of $\mathbf{D}^{-1/2}\mathbf{W}\mathbf{D}^{-1/2}$ by computing the singular value decomposition \cite{Martinson_SVD} of $\widetilde{\mathbf{C}}\mathbf{D}^{-1/2}\in\mathbb{R}^{m\times n}$. To see this, let $\mathbf{U}\boldsymbol{\Sigma}\mathbf{P}^T$ be the singular value decomposition of $\widetilde{\mathbf{C}}\mathbf{D}^{-1/2}$, where $\mathbf{U}\in\mathbb{R}^{m\times r}$ and $\mathbf{P}\in\mathbb{R}^{n\times r}$ have orthonormal columns, and $r$ is the rank parameter. Then, we have 
\begin{align}
\mathbf{D}^{-1/2}\mathbf{W}\mathbf{D}^{-1/2} = (\mathbf{U}\boldsymbol{\Sigma}\mathbf{P}^T)^T(\mathbf{U}\boldsymbol{\Sigma}\mathbf{P}^T)=\mathbf{P}\boldsymbol{\Sigma}^2\mathbf{P}^T.
\end{align}
Therefore, the top  $K$ eigenvectors of $\mathbf{D}^{-1/2}\mathbf{W}\mathbf{D}^{-1/2}$ are the same as the top $K$ right singular vectors of $\widetilde{\mathbf{C}}\mathbf{D}^{-1/2}$. As a result, the computational complexity of spectral clustering in our framework is linear with respect to the size of data $n$, which makes it suitable for large-scale datasets. The proposed method for SSC using landmarks is summarized in Alg.~\ref{alg:StandardNys}.

\begin{algorithm}[h]
	\caption{Fast Subspace Clustering (FSC)}
	\label{alg:StandardNys}
	\textbf{Input:}  Data $\mathcal{X}$,  number of landmarks $m$, parameter $\lambda$, number of subspaces $K$ 
	
	\textbf{Output:} Segmentation of   $\mathcal{X}$
	\begin{algorithmic}[1]
		\STATE Select $m$ landmarks $\mathcal{X}_0$ using uniform sampling or nonuniform sampling (K-medoids clustering)
		\STATE Solve the optimization problem in \eqref{eq:our} to find $\mathbf{C}\in\mathbb{R}^{m\times n}$
		\STATE $\widetilde{\mathbf{C}}=|\mathbf{C}|$, $\boldsymbol{\alpha}=\sum_{j=1}^n\widetilde{\mathbf{c}}_j$
		\STATE $\mathbf{D}=\text{diag}(\widetilde{\mathbf{C}}^T\boldsymbol{\alpha})$
		\STATE Run the K-means clustering algorithm on the top $K$ right singular vectors of $\widetilde{\mathbf{C}}\mathbf{D}^{-1/2}=\mathbf{U}\boldsymbol{\Sigma}\mathbf{P}^T$
	\end{algorithmic}
\end{algorithm}

\section{Experimental Results}
\label{sec:illust}
In this section, we demonstrate the effectiveness of our method fast subspace clustering for large-scale SSC. To this end,  the performance and time complexity of our method based on uniform sampling (FSC-uniform) and  nonuniform sampling using K-medoids clustering (FSC-nonuniform) are compared with the related work discussed in  Section \ref{sec:prior}. Recall that SSC-ADMM \cite{elhamifar2013sparse}, its modified version \cite{pourkamali2018efficient}, and SSC-OMP \cite{you2016scalable} solve the SSC problem using the full dataset without any landmark selection. To implement these algorithms, we use the code provided by the respective authors for computing the segmentation of data, and the parameters are selected to achieve the best clustering accuracy.

On the other hand, our proposed method and ESC \cite{you2018scalable} choose a set of representative points of size $m$ to solve the reduced SSC problem. The  optimization problems in \eqref{eq:EX}
and \eqref{eq:our} are solved using the SPAMS (SPArse Modeling Software) package \cite{mairal2014sparse}. To perform K-means and K-medoids clustering,  we use MATLAB's built-in functions with the maximum number of iterations is set to be 100; the City Block (Manhattan) distance is used for K-medoids clustering. 

The metric we use in this paper to compare the performance of   SSC methods is the clustering accuracy, which measures the maximum proportion of data points that are correctly labeled over all possible permutations of  the returned labels \cite{you2018scalable}. Hence, the clustering accuracy falls between 0 and 1, and values closer to 1 indicate better performance. 

In the first experiment, we consider the statistical model explained in \eqref{eq:data}, i.e., $\mathbf{x}_i=\mathbf{H}_k\mathbf{z}_i+\mathbf{e}_i$. This model allows us to control the ambient dimension $D$, the number of subspaces $K$, their dimensions, and the number of points per each subspace. We set parameters $D=16$, $K=5$, and $d_k=6$ for $k=1,\ldots,5$. The columns of the orthonormal matrices $\mathbf{H}_k\in\mathbb{R}^{D\times d_k}$ are drawn uniformly at random from a set of $D$ orthonormal random vectors in $\mathbb{R}^D$. The noise vectors $\mathbf{e}_i\in\mathbb{R}^D$, $i=1,\ldots,n$, are drawn i.i.d.~according to the Gaussian distribution $\mathcal{N}(0,\sigma^2)$, where $\sigma=0.1$. 

\begin{figure*}[t] 
	
	\begin{minipage}[b]{0.49\linewidth}
		\centering
		\centerline{\includegraphics[width=7.8cm]{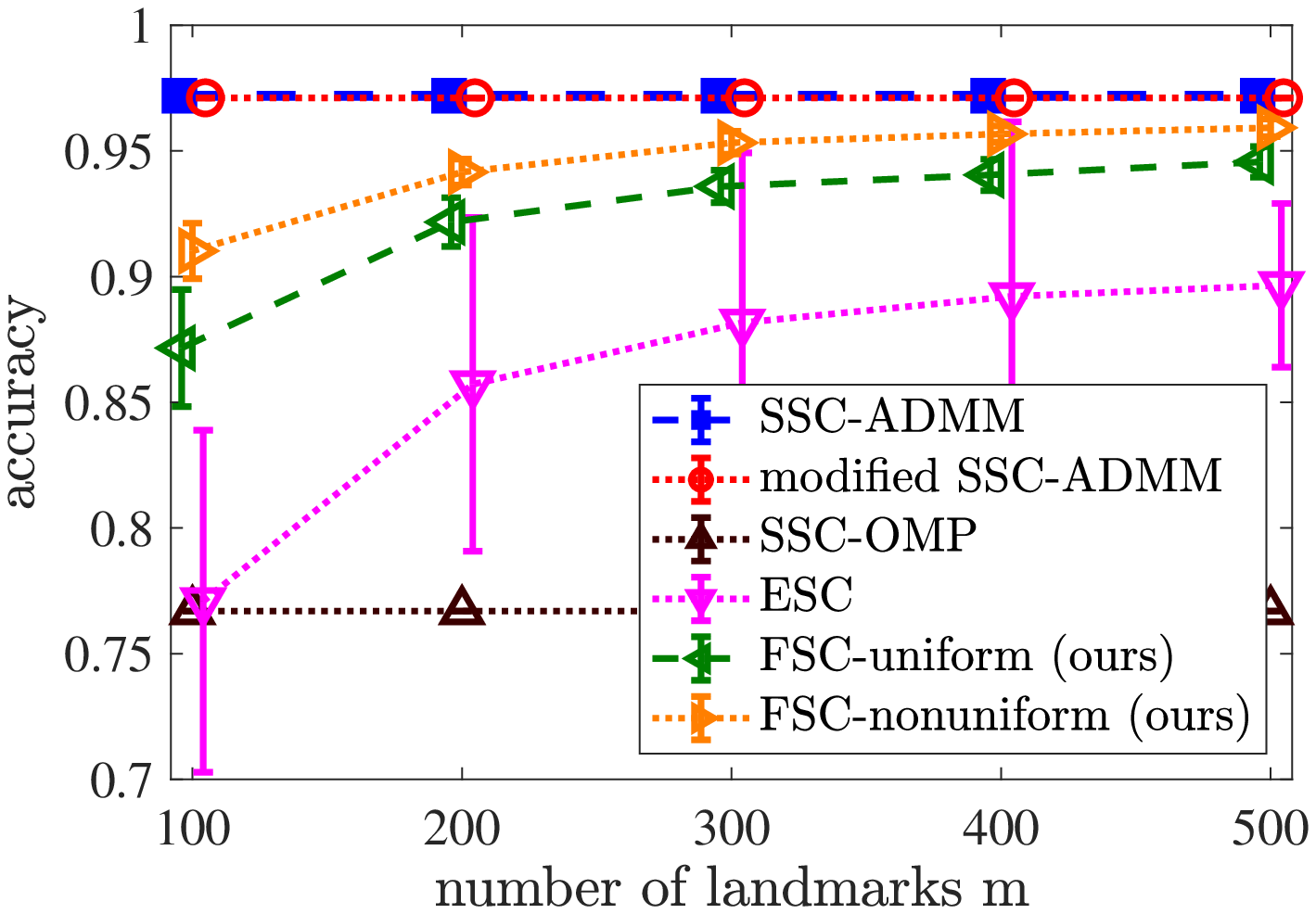}}
		\centerline{{(a) clustering accuracy, fixed $n=3,\!600$}}
	\end{minipage}
	\begin{minipage}[b]{0.49\linewidth}
		\centering
		\centerline{\includegraphics[width=7.8cm]{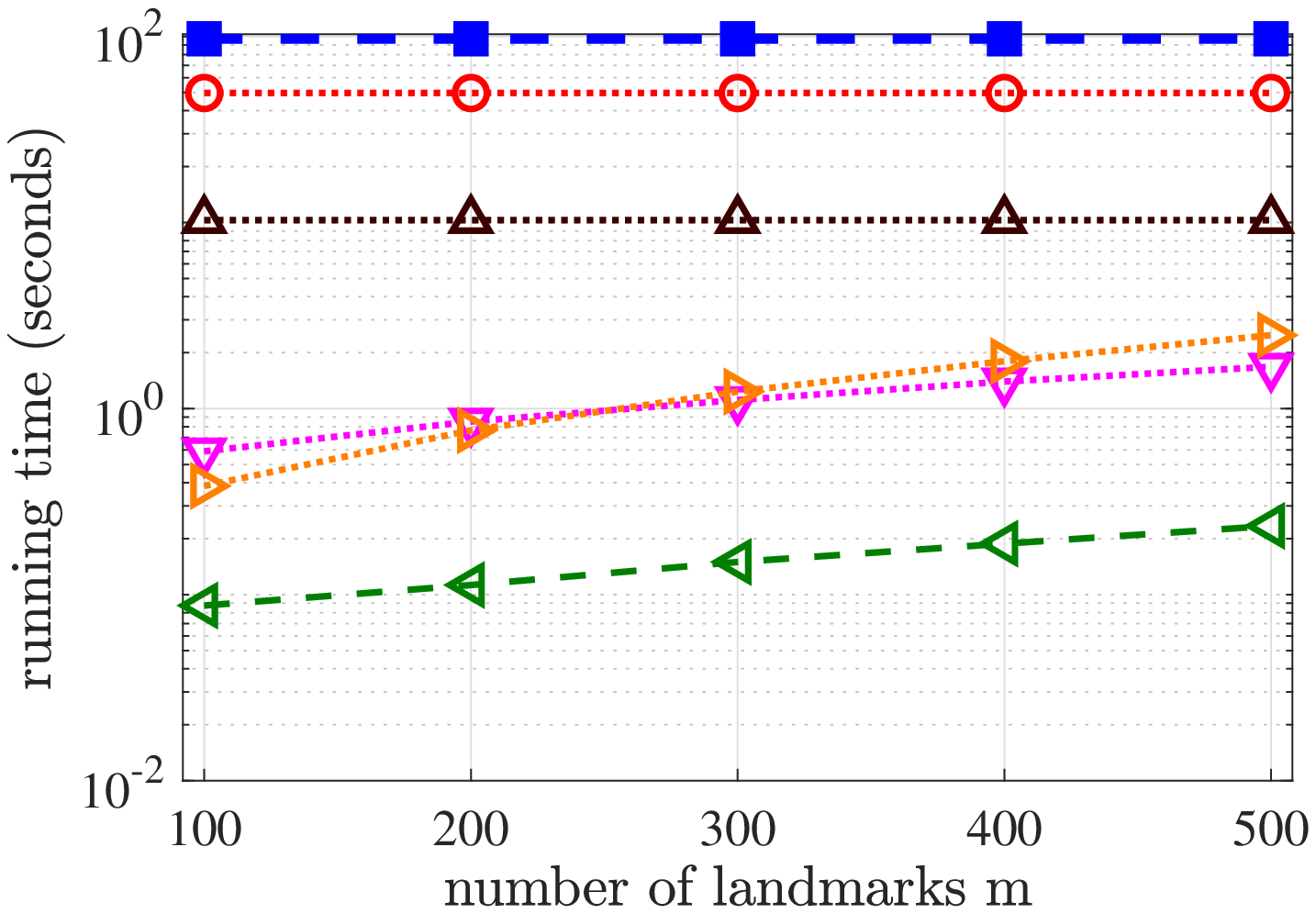}}
		\centerline{{(b) runtime, fixed $n=3,\!600$}}
	\end{minipage}
	
	\medskip
	\medskip
	\begin{minipage}[b]{0.49\linewidth}
		\centering
		\centerline{\includegraphics[width=7.8cm]{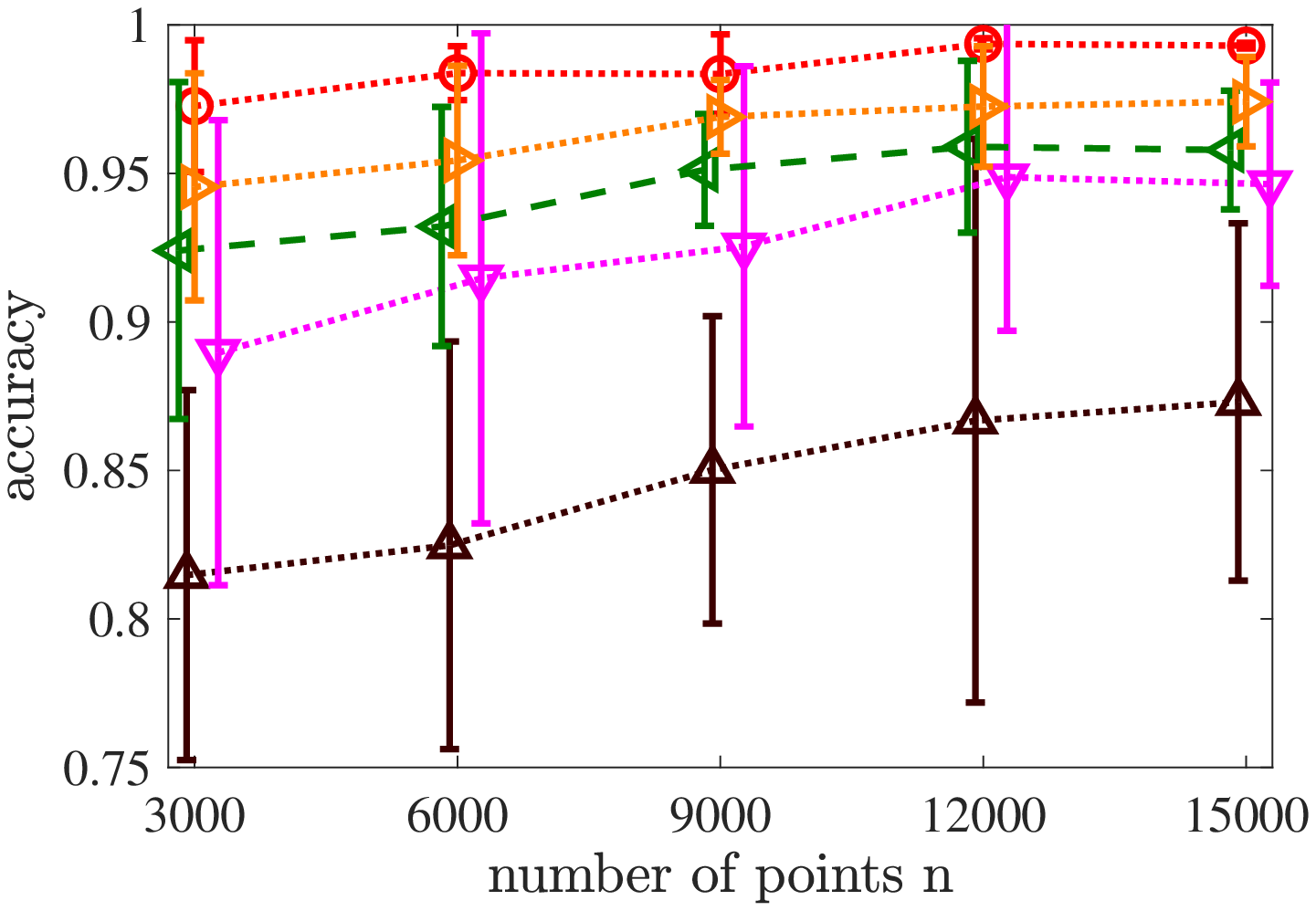}}
		\centerline{{(c) clustering accuracy, fixed $m=300$}}
	\end{minipage}
	\begin{minipage}[b]{0.49\linewidth}
		\centering
		\centerline{\includegraphics[width=7.8cm]{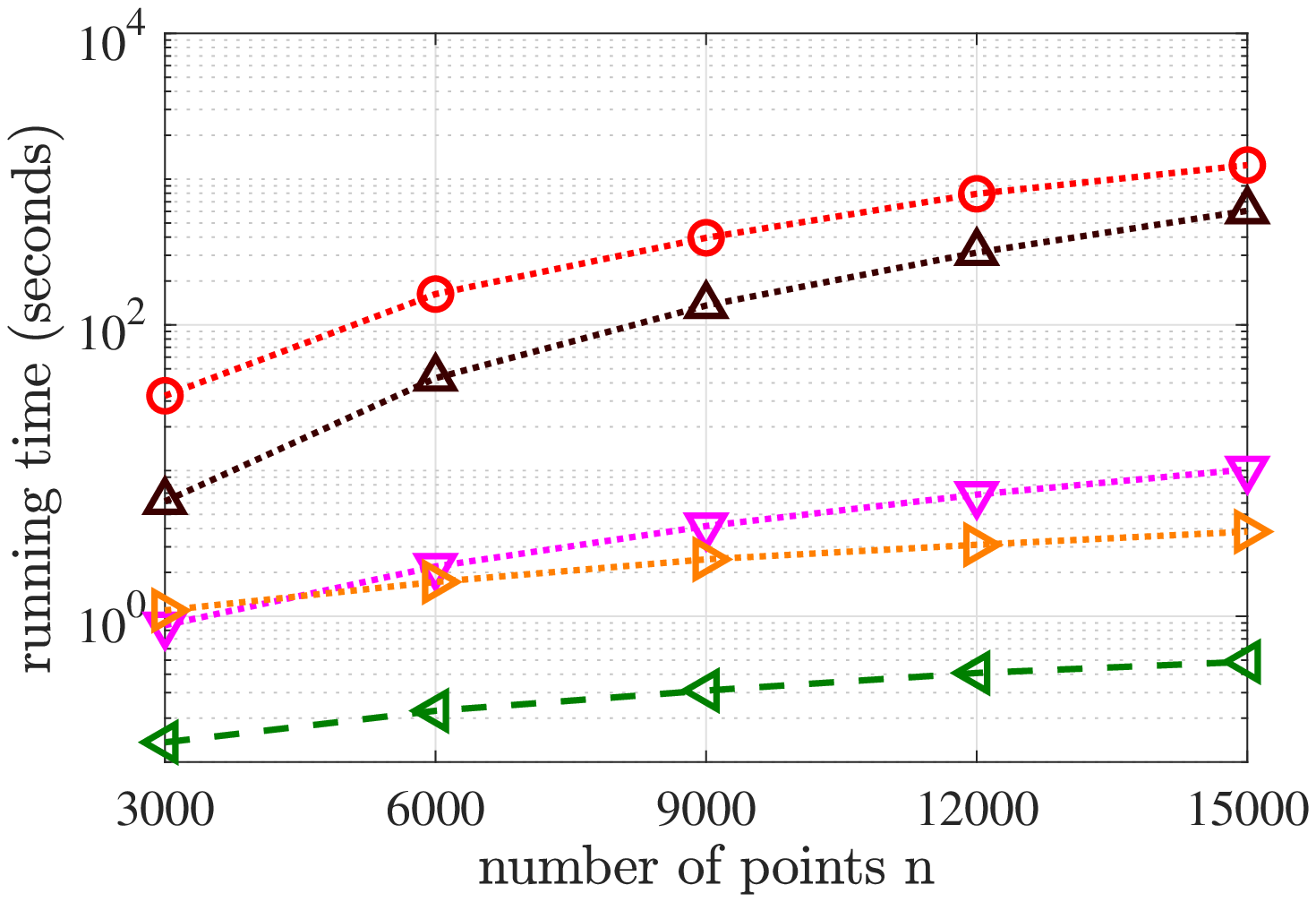}}
		\centerline{{(d) runtime, fixed $m=300$}}
	\end{minipage}
	\hfill
	\caption{Comparing the performance and efficiency of our  fast subspace clustering (FSC) with the related work on  synthetic datasets. Note that the legend in (a) refers to all other figures in (b), (c), and (d). }
	\label{fig:res}
\end{figure*}

To demonstrate the impact of landmarks, we set the number of  points per subspace to be $720$, which means that the total number of input data points is $n=3,\!600$. Fig.~\ref{fig:res}(a) shows the mean and standard deviation of clustering accuracy over $20$  trials for varying number of landmarks $m$ in the range of $100$ to $500$. Moreover, the averaged runtime of SSC methods is reported in Fig.~\ref{fig:res}(b). As expected, both SSC-ADMM and its modified version  perform equally well, while the latter is almost two times faster in this experiment. We also see that SSC-OMP is more efficient than SSC-ADMM and its modified version (an order of magnitude faster than ADMM). However,  SSC-OMP is less accurate compared to ADMM which is consistent with prior observations, e.g., \cite{you2016scalable}. In contrast, our proposed FSC-uniform significantly outperforms SSC-OMP and ESC, while it is faster than the other methods by a large margin. In fact, FSC-uniform reaches $90\%$ accuracy  when the number of landmarks is $200$. Furthermore, the nonuniform sampling scheme based on K-medoids clustering is slightly more accurate than FSC-uniform, showing the importance of landmark selection specifically for small values of $m$. Although our FSC-nonuniform and ESC run in roughly the same amount of time,  our method is more accurate than ESC for varying $m$.

In the next experiment, we consider the same synthetic dataset with a fixed number of landmarks $m=300$ and varying number of data points from $n=3,\!000$ to $n=15,\!000$. The mean and standard deviation of clustering accuracy over $20$ trials are reported in Fig.~\ref{fig:res}(c), and we report the averaged runtime results in Fig.~\ref{fig:res}(d). As we observed in the previous experiment, SSC-ADMM is slower than its modified version so we dropped this method to make the figures easier to read. We see that both versions of our FSC outperform SSC-OMP and ESC, and they almost reach the accuracy of solving the sparse optimization using the entire dataset via ADMM. In terms of running time, we see that our FSC-uniform is faster than the other methods by a large margin and it scales linearly as the number of data points $n$ increases (the average runtime of FSC-uniform for $n=15,\!000$ is about $0.45$ seconds). Thus, the proposed method is suitable for performing SSC on   large-scale datasets. In addition, the computational cost of FSC-nonuniform increases at a lower rate compared to ESC as the number of data points $n$ increases.

In the last experiment, we evaluate the performance of SSC methods on clustering images of handwritten digits. To this end, we use the MNIST dataset which contains gray-scale images \cite{lecun1998gradient}. According to \cite{you2016scalable}, we compute a set of nonlinear features using a scattering convolutional neural network \cite{bruna2013invariant}. Then, we form a feature vector for each original image  of size $28\times 28$ by concatenating coefficients in each layer of the network.  The dimensionality of the resulting feature vector for each image is $3,\!472$, and the authors in \cite{you2016scalable} proposed to reduce the dimensionality of these feature vectors using principal component analysis (PCA) from $3,\!472$ to $500$. To follow the same path, we also perform PCA on the feature vectors. However, one can instead use randomized dimension reduction methods  that are shown to be useful for achieving high accuracy in the SSC framework \cite{Pourkamali_ICML_2014,heckel2017dimensionality,meng2018general}.

\begin{figure}[h] 
	
	\begin{minipage}[b]{0.98\linewidth}
		\centering
		\centerline{\includegraphics[width=8.5cm]{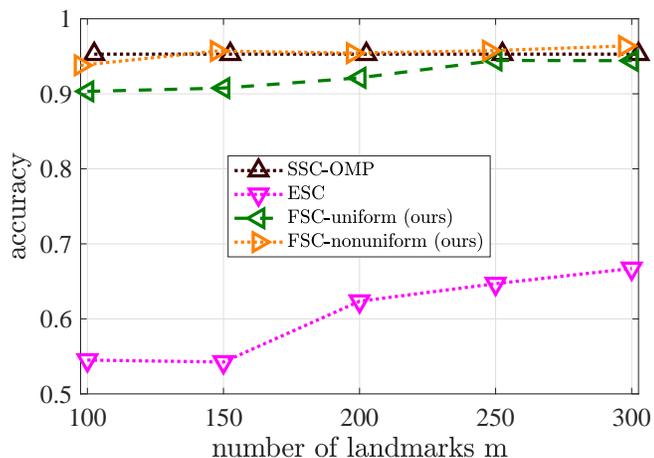}}
	\end{minipage}
	
	\hfill
	\caption{SSC on the MNIST dataset for fixed $n=8,\!000$.}
	\label{fig:res2}
\end{figure}

In summary, we consider $2,\!000$ images randomly chosen from 4 clusters of digits  2, 7, 8, and 9. Thus, we have a dataset with the ambient dimension $D=500$ and the total number of points $n=8,\!000$, which consists of $K=4$ clusters. Since this dataset has been shown to be well-suited for SSC-OMP \cite{you2016scalable}, we present the averaged clustering accuracy of landmark-based SSC methods for varying values of $m$ in Fig.~\ref{fig:res2} . As we see, our proposed methods outperform ESC by a large margin for all values of landmarks. For small values of $m$, our nonuniform sampling scheme is slightly more accurate than uniform sampling, which again indicates the importance of landmark selection when $m$ is small compared to $n$. However, as the number of landmarks increases, the difference between uniform and nonuniform sampling becomes less significant.

\section{Conclusion}
\label{sec:conc}
In this paper, we presented an efficient approach to subspace clustering on large-scale datasets using landmarks. Two sampling schemes are used to select a small number of landmarks for solving the sparse optimization problem.  Then, we showed that the resulting coefficient matrix can be directly used for performing spectral clustering to achieve the data segmentation. Thus, our method provides tunable tradeoffs between clustering accuracy and time complexity. Empirically, we observed a significant reduction in computational time, while achieving accurate clustering results competitive with the models trained on the full dataset.

\bibliographystyle{ieeetr}
\bibliography{phd_farhad}

\end{document}